\documentclass[11pt,a4paper]{article}
\usepackage[margin=25mm]{geometry}
\usepackage[T1]{fontenc}
\usepackage{lmodern}
\usepackage{textcomp,microtype}
\usepackage{amsmath,amssymb}
\usepackage{graphicx}
\usepackage{booktabs,tabularx,array}
\usepackage[numbers,sort&compress]{natbib}
\usepackage[hidelinks]{hyperref}
\newcommand{\code}[1]{\texttt{#1}}
\newcommand{\KeepWithFollowing}[1]{%
  \par\begingroup
  \dimen0=\pagegoal\advance\dimen0 by -\pagetotal
  \ifdim\dimen0<#1\newpage\fi
  \endgroup
}
\hypersetup{
  pdftitle={Recoverability as a System Primitive for Long-Horizon AI Agents},
  pdfauthor={Zhihui Zhang; Wei Liu},
  pdfsubject={Research preprint}
}
\title{Recoverability as a System Primitive for Long-Horizon AI Agents}
\author{%
  Zhihui Zhang\thanks{Independent researcher; corresponding author.
  Email: \href{mailto:zhangzhmail@126.com}{zhangzhmail@126.com}.}
  \and
  Wei Liu\thanks{Beijing Information Science and Technology University.
  Email: \href{mailto:willie@bistu.edu.cn}{willie@bistu.edu.cn}.}
}
\date{}
\begin{document}
\maketitle
\begin{abstract}
AI agents can be interrupted while editing files, calling tools, or carrying out multi-\hspace{0pt}step tasks. Restarting repeats completed work, but continuing from unverified or outdated progress can carry earlier errors forward. A saved state is not necessarily a suitable place to resume. We introduce \textbf{recoverability} as a system primitive that makes reuse an explicit decision: select a supported starting point and a permitted recovery action, or withhold automatic continuation. Its behavioral contract binds that choice to supporting evidence, execution, and independent checks. A reference architecture connects persistence, validation, and control, with complementary runtime instances testing distinct responsibilities. Four deterministic and 20 paired file challenges demonstrate that accurate restoration and successful completion can conceal disallowed starting points. Progress controls attribute retained work to shared restoration. Event-\hspace{0pt}time tests show that permission must also constrain the action, and that independently held policy evidence can expose violations even after an effect occurs. These findings establish why recovery decisions need their own evaluation, beyond restored bytes and final task success. Within supplied policies and a declared trust model, the contribution is a common, testable interface for retaining justified progress and making the conditions for its reuse explicit and enforceable.
\end{abstract}

\section{Introduction}
An AI agent may spend many steps editing a document, transforming a table, or calling tools before its execution is interrupted. It may also keep running while repeating an ineffective action or failing to make validated progress. Recovery then presents a practical choice. Starting over repeats work that has already been checked, while continuing from the latest state may reuse work that failed verification or is no longer supported by current evidence. The system needs to determine which progress can still be used and how to proceed from it.

Consider a document-\hspace{0pt}editing task. Checkpoint A contains an approved version. The agent later saves an exploratory edit at checkpoint B, but the edit fails a required check. A subsequent tool call fails. Restarting the whole task repeats the approved work in A; restoring the newest checkpoint selects B despite its failed check. If A's approval still applies and retry is permitted, the agent can instead resume from A. The useful distinction is simple: \textbf{a state being saved does not mean that it remains suitable for continuation.}

\begin{figure}[htbp]
  \centering
  \includegraphics[width=\linewidth]{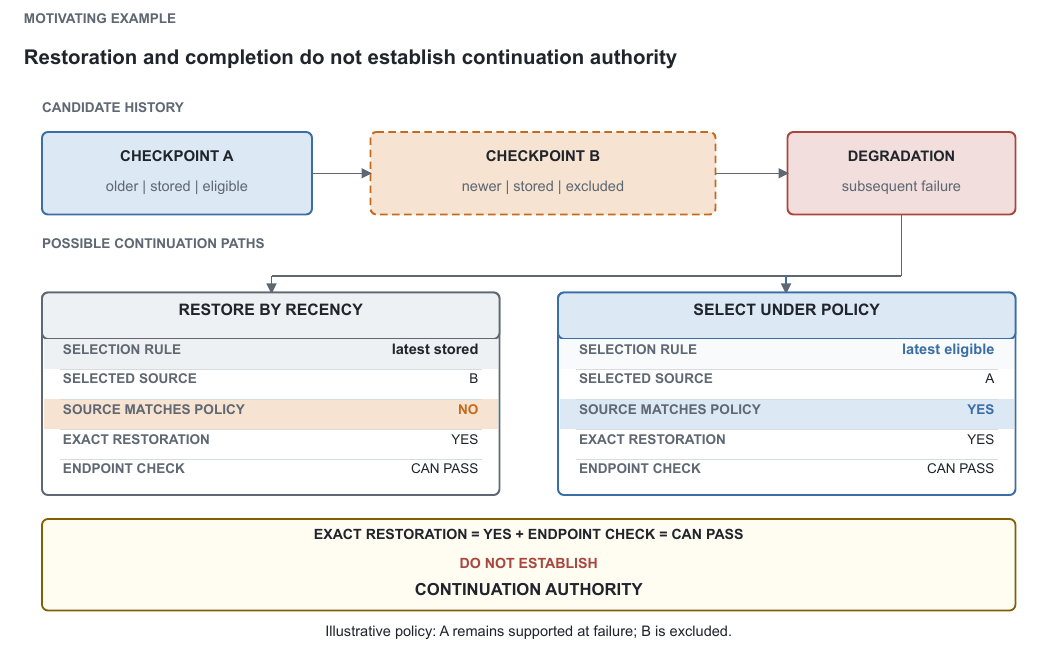}
  \caption{Choosing where to resume. A passed the required check and remains eligible at failure; newer B is excluded. Either snapshot can be restored exactly and later repaired to a valid document. Those outcomes do not reveal whether the chosen starting point met the continuation conditions.}
  \label{fig:1:continuation_boundary_control}
\end{figure}
Choosing a recovery point is only part of the decision. Retrying a read may be appropriate, while retrying a tool call that might already have changed external state could duplicate an operation. An old approval may also have expired. Even a previously validated checkpoint may therefore provide no justified way to continue automatically. The system must consider the saved state together with the next recovery action, and be able to stop or request review when their conditions are not met.

The central question is:

\begin{quote}
After an agent fails or loses reliable progress, what work can it retain, where should it resume, and what should it do next?
\end{quote}

Checkpointing provides saved states and restoration mechanisms. Validators supply evidence about particular versions, diagnosis identifies what went wrong, and a planner or workflow chooses possible corrective actions. These mechanisms are essential, but their outputs must be connected: a stored version needs evidence supporting its reuse, and a proposed action needs conditions under which it may run. A retry rule alone need not check either relation. A checkpoint manager or workflow engine can provide this control if it explicitly decides where and how execution is allowed to continue.

We call this responsibility \textbf{recoverability}: the system's ability to select an evidence-\hspace{0pt}supported recovery point and an allowed recovery action after degradation, or to withhold automatic continuation when the governing rules require it. We treat it as a \emph{system primitive} because it has its own inputs, decision, and requirements that other components can implement. It coordinates checkpointing, validation, and repair through the execution control plane. It does not generate the repair content itself.

For precision, we call the selected state a \textbf{continuation source}, the corrective action a \textbf{recovery route}, and permission to execute that particular combination \textbf{continuation authority}. Authority is granted under evidence and policy at the failure event; it is not a permanent label on a checkpoint. The proposed contract requires the decision to govern execution, remain checkable afterward, and respect limits on repeated recovery.

This makes recovery quality visible beyond the final task result. In the document example, an agent might restore B and later repair it successfully. The final document would then conceal the earlier choice of a recovery point that the policy excluded. Checking whether the saved bytes were restored exactly would miss the same error. The source, the action, and the permission to execute must therefore be evaluated alongside restoration and completion.

\KeepWithFollowing{7\baselineskip}
This paper makes three contributions:

\begin{enumerate}
  \item \textbf{A testable contract for reusing progress.} We formalize which saved state and recovery action may be used together, including required continuation and required withholding. Four requirements connect this decision to its supporting evidence, execution, and bounds. They make recovery correctness separately checkable from restoration fidelity and task completion, including when those observations conceal a disallowed choice.
  \item \textbf{A reference architecture and bounded implementations.} We organize candidate management, evidence assessment, decision, execution, and audit into an explicit control process. The architecture connects existing checkpointing and validation mechanisms to permission for a particular recovery transition. Complementary implementations make source selection, restoration, and event-\hspace{0pt}time permission concrete, with their supported behaviors and assumptions stated separately.
  \item \textbf{Controlled evidence for evaluating recovery decisions.} Source-\hspace{0pt}choice challenges demonstrate that accurate restoration and successful endpoints can conceal a disallowed starting point. Progress controls attribute retained work to shared checkpoint restoration. Authorization tests distinguish enforcing a decision from independently checking the policy under which it was executed. Together, these results establish observable recovery errors that task success alone cannot reveal.
\end{enumerate}

The aim is to retain work whose reuse is justified and make the conditions for continuing explicit. The experiments use supplied validators and policies, so they test conformance to those conditions rather than discover trust in an open environment. Section 3 defines the common contract, Sections 4 and 5 explain its design and bounded implementations, and Section 6 examines what each study establishes.

\section{Related Work}
Recovery involves saving work, assessing its validity, deciding what to do next, and executing that decision. These responsibilities appear in several established research areas. We position recoverability by asking how each contributes to selecting a justified place and action from which to continue.

\subsection{Fault Tolerance and Checkpointing}
Distributed snapshots capture consistent global states \cite{chandy1985distributed}, and rollback-\hspace{0pt}recovery protocols explain how computation can resume from consistent checkpoints after failure \cite{elnozahy2002survey}. They provide foundations for preserving and restoring work. For an agent, a computationally consistent snapshot can still contain a document that failed validation or a plan whose assumptions have become outdated. The restoration mechanism needs a decision about whether that state remains suitable for use.

Agent checkpointing increasingly accounts for state outside the conversation. Crab uses effect-\hspace{0pt}aware checkpointing to capture operating-\hspace{0pt}system effects in agent sandboxes and improve restoration correctness \cite{wu2026crab}. Its focus is what a checkpoint must capture and when capture is useful. Recoverability examines the choice among available states and the actions allowed from them. A more complete snapshot helps execute that choice accurately; it does not alone establish that the chosen version meets current continuation conditions.

The closest alternative is a checkpoint manager with semantic guards. Filtering candidates by validation already implements part of the responsibility. If the manager also checks whether an action is permitted from the selected state, supports required refusal, enforces the decision, and records its rationale and bounds, it may satisfy the full recoverability contract. Our contribution is specifying this combination as an observable recovery decision and evaluating its validity separately from restoration and task outcomes. Restoring a rejected version and repairing it successfully remains an incorrect recovery choice under a policy that excluded it. This criterion can be applied to an existing checkpoint manager without claiming exclusive ownership of its components. The restore-\hspace{0pt}latest baseline in our experiments is one deliberately explicit selection policy, not a characterization of all checkpoint systems.

\subsection{Memory and Persistent Agent State}
Reflexion uses linguistic feedback to improve later attempts \cite{shinn2023reflexion}, Voyager retains executable skills during exploration \cite{wang2023voyager}, and Generative Agents retrieve and synthesize stored experiences \cite{park2023generative}. MemGPT organizes multiple memory tiers for extended language-\hspace{0pt}model interactions \cite{packer2023memgpt}. These approaches make information persist beyond a single context.

Remembering a task is useful, but does not establish which version should be edited next. A retained claim may refer to an older artifact or conflict with a later validation result. In the recoverability design, memory can inform the choice of recovery point, while its relevance must be checked against the current artifacts and evidence. The present runtimes do not evaluate memory-\hspace{0pt}conflict resolution.

\subsection{Diagnosis, Adaptation, and Workflow Control}
ReAct interleaves reasoning and action \cite{yao2023react}, while Toolformer expands model use of external tools \cite{schick2023toolformer}. Workflow systems organize dependencies, retries, fallbacks, and persistent execution. Observability supplies the traces needed to inspect these behaviors.

Research on agent recovery also studies how to choose better actions after failure. \emph{Hell or High Water} examines alternative planning when previously available functions fail \cite{wang2025hell}. AgentDebug localizes errors in modular trajectories and supplies targeted corrections \cite{zhu2025where}. These approaches address diagnosis and adaptation. Before a correction executes, however, the system still needs to determine which persistent state supports it and what effects are permissible from that state.

Recoverability connects these outputs to the state being reused. For example, a diagnosis of a failed tool call does not by itself determine whether the call changed a file before failing, which snapshot should now be selected, or whether retry may duplicate the effect. A workflow can answer these questions through its own policy and implement the proposed contract. The contract makes the answers explicit and checkable without requiring a particular module layout.

\subsection{Evaluating Agent Execution}
SWE-\hspace{0pt}bench measures issue resolution in software repositories \cite{jimenez2024swebench}, WebArena evaluates web interaction \cite{zhou2024webarena}, OSWorld evaluates computer use \cite{xie2024osworld}, and GAIA tests general assistant capabilities requiring tools and reasoning \cite{mialon2024gaia}. ToolSandbox adds stateful tool execution, implicit dependencies, and intermediate milestones \cite{lu2025toolsandbox}. Agent evaluation consequently encompasses both outcomes and execution behavior \cite{mohammadi2025evaluation}.

Our evaluation asks whether recovery used an allowed starting point and action, as well as whether the task finished. An endpoint score alone cannot show that validated work was discarded or that execution resumed from a rejected version. We therefore record source selection, restoration, action counts, permission and its execution, and final invariants separately. The studies are controlled demonstrations of these distinctions, not a workload-\hspace{0pt}scale comparison of agent intelligence or deployed reliability.

This positioning is behavioral. A semantic checkpoint manager or workflow engine may already meet some or all of the proposed requirements. Determining that requires examining its decision and enforcement behavior, including whether the basis for accepting an execution remains independent of a client that can alter the executed policy. The current experiments isolate selected requirements over shared mechanisms; a direct empirical comparison with such complete systems remains outside their scope.

\section{Problem Formulation}
The running example raises three questions: which saved state can be reused, which action is allowed from it, and whether execution should continue at all. We now express these questions independently of a particular runtime. The formulation distinguishes the decision from two later observations: whether the selected state was restored accurately and whether the task eventually succeeded.

\subsection{Execution State and Degradation}
Consider an execution trajectory

\begin{equation*}
\tau=(o_0,a_0,o_1,a_1,\ldots,o_T),
\end{equation*}

where observations include instructions, tool results, artifact states, and validation signals, and actions include model outputs, tool calls, edits, and verification requests. Because actions can change persistent state, conversational context is only one part of the execution. Write

\begin{equation*}
X_t=(S_t,A_t,M_t,H_t,C_t,V_t),
\end{equation*}

where $S_t$ is operational state, $A_t$ artifact versions, $M_t$ retained claims, $H_t$ execution history, $C_t$ stored checkpoints, and $V_t$ validation records. Implementations may omit or extend these components. The model permits disagreement among them: a memory may describe an older artifact, a checkpoint may contain unvalidated work, or a trace may record a call whose external effect remains uncertain.

A state is \textbf{degraded} when execution can no longer proceed under its usual assumptions without checking them again. Let $f_t$ denote the detected event. An explicit error is one example; others include repeated action without validated progress, stale evidence, artifact corruption, incomplete context, and uncertainty about a tool's side effects. Detection triggers a recovery decision. It does not itself establish where execution should resume or whether the next action is allowed.

These examples describe situations in which recovery may be needed. A detector must establish the relevant trigger within its own scope; the formulation does not assume that every kind of degradation can be recognized by one mechanism.

\subsection{Recovery Points and Conditions for Reuse}
We use \textbf{candidate continuation boundary} for a possible recovery point: a state reference together with the information needed to continue from it. It may include an artifact version, a checkpoint, an execution cursor, and relevant context. Let $B_t$ be the available candidates, $\mathcal{E}_t$ the available evidence, and $\Pi_t$ the governing continuation policy. To distinguish a stored candidate from one that meets the current conditions for reuse, define

\begin{equation*}
\operatorname{Supported}(b\mid X_t,f_t,\mathcal{E}_t,\Pi_t)
\end{equation*}

to mean that the evidence and policy support using $b$ as a source after event $f_t$. This is what \textbf{eligible} means in the model. Support may depend on validation, provenance, freshness, uncertainty, and human approval. The predicate specifies the judgment a runtime must obtain; it does not supply a universal method for determining trust.

Given a strict recency order $\prec$, or a declared deterministic tie-\hspace{0pt}break, define

\begin{equation*}
s_t=\max_{\prec}B_t,\qquad \ell_t=\max_{\prec}\{b\in B_t:\operatorname{Supported}(b\mid X_t,f_t,\mathcal{E}_t,\Pi_t)\}.
\end{equation*}

Either value is $\varnothing$ if its set is empty. Thus $s_t$ is the newest available state, whereas $\ell_t$ is the newest state supported for reuse at this event. In the document example, these can be B and A, respectively. A's earlier approval is relevant evidence, but using it later also requires that the approval still applies. Eligibility depends on the event, evidence, and rules; storing a label on a checkpoint does not make that label permanently valid.

\subsection{Recovery Actions and Permission to Continue}
A recovery point does not determine what to do next. A \textbf{recovery route} specifies that action or disposition: retrying, replanning, restoring earlier state, or stopping automatic continuation. Separate routes that continue task execution from those that withhold it:

\begin{equation*}
\mathcal{R}=\mathcal{R}_{\mathrm{cont}}\mathbin{\dot\cup}\mathcal{R}_{\mathrm{withhold}}.
\end{equation*}

Continuing routes include retry, rollback, resume, and replan. Withholding routes include halt or escalation without automatic continuation. The governing profile must state whether a compound action such as freeze-\hspace{0pt}and-\hspace{0pt}verify permits continued task execution or only obtains evidence for a later decision.

The route should reflect the failure and its possible effects. A reversible execution error may permit retry; repeated ineffective action may call for replanning; loss of validated progress may require rollback followed by replan. These are conditional examples, not a routing rule inferred from the failure label alone.

For a continuing route, both the state and the action must meet their conditions. Let $\mathcal{R}_{\mathrm{valid}}$ contain the routes allowed for the given event, source, evidence, and policy. The permitted source-\hspace{0pt}and-\hspace{0pt}route pairs are

\begin{equation*}
\begin{aligned}
\mathcal{P}_t=\{(b,r):{}& b\in B_t,\\
& \operatorname{Supported}(b\mid X_t,f_t,\mathcal{E}_t,\Pi_t),\\
& r\in\mathcal{R}_{\mathrm{cont}}\cap{}\mathcal{R}_{\mathrm{valid}}
(f_t,b,X_t,\mathcal{E}_t,\Pi_t)\}.
\end{aligned}
\end{equation*}

An eligible source can have no valid continuing route. For example, A may still pass document validation while an uncertain external write makes a retry unacceptable. Conversely, a plausible repair does not justify starting from a version that the policy excludes. A runtime can search for source and route together or reconsider one after checking the other; the model does not prescribe their computation order.

The runtime then issues \textbf{continuation authority}: permission for a particular source-\hspace{0pt}and-\hspace{0pt}route transition, or a decision to withhold automatic continuation. Write

\begin{equation*}
\rho(X_t,f_t,\mathcal{E}_t,\Pi_t)\rightarrow(g_t,b_t^*,r_t),\qquad g_t\in\{\mathrm{grant},\mathrm{withhold}\}.
\end{equation*}

The allowed decisions include permitted continuing pairs and withholding dispositions that meet their own policy conditions:

\begin{equation*}
\begin{aligned}
\mathcal{D}_t={}&\{(\mathrm{grant},b,r):(b,r)\in\mathcal{P}_t\}\;\cup\\
&\{(\mathrm{withhold},\varnothing,r):
r\in\mathcal{R}_{\mathrm{withhold}}\cap{}\mathcal{R}^{\mathrm{withhold}}_{\mathrm{valid}}
(f_t,X_t,\mathcal{E}_t,\Pi_t)\}.
\end{aligned}
\end{equation*}

A decision is policy-\hspace{0pt}correct when it belongs to $\mathcal{D}_t$. A \textbf{declared profile} supplies the admission, action, and withholding rules before the controller is evaluated. It is well formed only when $\mathcal{D}_t\ne\varnothing$ for every in-\hspace{0pt}scope event. Having an eligible candidate $\ell_t$ does not force continuation: an action may be unavailable or withholding may be permitted. The granted source $b_t^*$ need not be the most recent eligible candidate if another source and route satisfy the policy.

We call $b_t^*$ a \textbf{trusted continuation boundary} when it is the source of a granted policy-\hspace{0pt}valid transition. This term refers to support under the stated evidence and policy, not objective world truth. The decision alone is also insufficient: a trustworthy continuation requires that execution actually follow the granted transition.

\subsection{Recovery Correctness and Task Completion}
After a decision, execution provides further observations. Let $p_t$ be the source actually selected. When restoration is used, $F(\hat p_t,p_t)$ reports whether reconstruction matches that source, and $Y$ reports whether the final artifact satisfies the task invariant. Neither observation answers whether the decision belonged to $\mathcal{D}_t$. The runtime can faithfully restore a rejected version, then repair it successfully. Table 1 separates the resulting combinations.

\begin{table}[htbp]
  \centering
  \small
  \renewcommand{\arraystretch}{1.12}
  \caption{Task outcomes and recovery decisions answer different questions.}
  \label{tab:1}
  \begin{tabularx}{\linewidth}{@{}>{\hsize=0.660\hsize\linewidth=\hsize\raggedright\arraybackslash}X>{\hsize=0.660\hsize\linewidth=\hsize\raggedright\arraybackslash}X>{\hsize=1.680\hsize\linewidth=\hsize\raggedright\arraybackslash}X@{}}
  \toprule
\textbf{Final invariant satisfied?} & \textbf{Decision policy-\hspace{0pt}correct?} & \textbf{Interpretation} \\
\midrule
Yes & Yes & A valid endpoint follows a policy-\hspace{0pt}correct decision \\
Yes & No & Endpoint success masks an incorrect recovery decision \\
No & Yes & A valid halt or escalation, or an unsuccessful granted continuation \\
No & No & Neither the decision nor the endpoint meets its requirement \\
  \bottomrule
  \end{tabularx}
\end{table}

The runtime's termination flag is another distinct observation: reaching a state named \code{completed} need not mean that the final invariant passed. Recoverability concerns making and executing a policy-\hspace{0pt}valid recovery decision, with separate evidence about the source, restoration, and outcome. It can therefore recognize a correct refusal as well as a correct continuation.

\subsection{Requirements for a Recoverable System}
The definition becomes a testable system contract through four requirements:

\begin{enumerate}
  \item \textbf{Distinguish saved state from eligible state.} Make candidates available without assuming that recency or restorability proves suitability for reuse.
  \item \textbf{Decide whether and how to continue.} Issue a source-\hspace{0pt}and-\hspace{0pt}route grant or withholding disposition in $\mathcal{D}_t$, with the reason supporting it.
  \item \textbf{Execute according to that decision.} Enact the authorized transition, check fidelity when restoring state, and stop automatic continuation when permission is withheld.
  \item \textbf{Retain evidence and limit recovery.} Preserve what is needed to check the decision and prevent unbounded repeated attempts.
\end{enumerate}

A full-\hspace{0pt}conformance evaluation must cover both directions. \textbf{Grant-\hspace{0pt}required} cases make withholding invalid; \textbf{withhold-\hspace{0pt}required} cases make every grant invalid. Otherwise a system that always stops could appear correct without ever performing useful recovery. This coverage is necessary within the declared scope and gives no guarantee beyond it.

A semantically guarded checkpoint manager or workflow engine can meet these requirements using its own components, with or without snapshot restoration. Partial implementations can also be useful when their tested responsibilities are explicit. The design below shows how to organize the decision; the following runtime and evaluation sections identify the parts actually exercised.

\section{Recoverability Primitive Design}
The design translates the contract in Section 3 into a runtime control path. It connects the work that can be retained, the evidence supporting its reuse, and the action that may follow. We first explain the design goals, then the reference architecture, its objects, and the lifecycle from ordinary execution through recovery and verification.

\subsection{Design Goals and Responsibility Boundaries}
Useful recovery depends on both retaining progress and preserving the conditions under which it can be reused. Saving an exploratory edit should not overwrite the record of an earlier approved version. Equally, an earlier approval should not silently apply to different content or remain valid after its supporting conditions have changed. The design therefore keeps candidate storage separate from evidence about candidate eligibility, with validation bound to specific versions and scopes.

The decision must cover both a recovery point and an action. A planner may propose a repair, a validator may accept a document, and a store may reconstruct its bytes, but these outputs still need to be connected to the conditions for continuing. Retry, rollback, resume, replan, verification, and escalation have different preconditions and effects. A suitable source cannot compensate for an invalid action, and a plausible action cannot justify an excluded source.

Recovery also needs an operational stopping rule. Withholding must prevent automatic continuation through the governed path, while permitted recovery must still be possible when the policy requires it. Attempts, execution, and retained state need bounds because recovery can itself repeat or stall. These goals motivate the four design principles shown in Figure 2: evidence-\hspace{0pt}grounded authorization, explicit eligibility and authority, failure-\hspace{0pt}specific routing, and bounded recovery.

\subsection{RPR Reference Control Architecture}
The \textbf{Recoverability Primitive Runtime (RPR)} is the reference design in Figure 2. It organizes the control path into an \textbf{Evidence/\hspace{0pt}Policy} layer, a \textbf{Governance} layer, and an \textbf{Enactment/\hspace{0pt}Audit} layer. The architecture specifies responsibilities and interfaces; a checkpoint manager, workflow engine, or agent runtime may combine modules or use a different persistence substrate. Its five stages express information and decision dependencies, not five separate processes or mutually exclusive time intervals.

\begin{figure}[htbp]
  \centering
  \includegraphics[width=\linewidth]{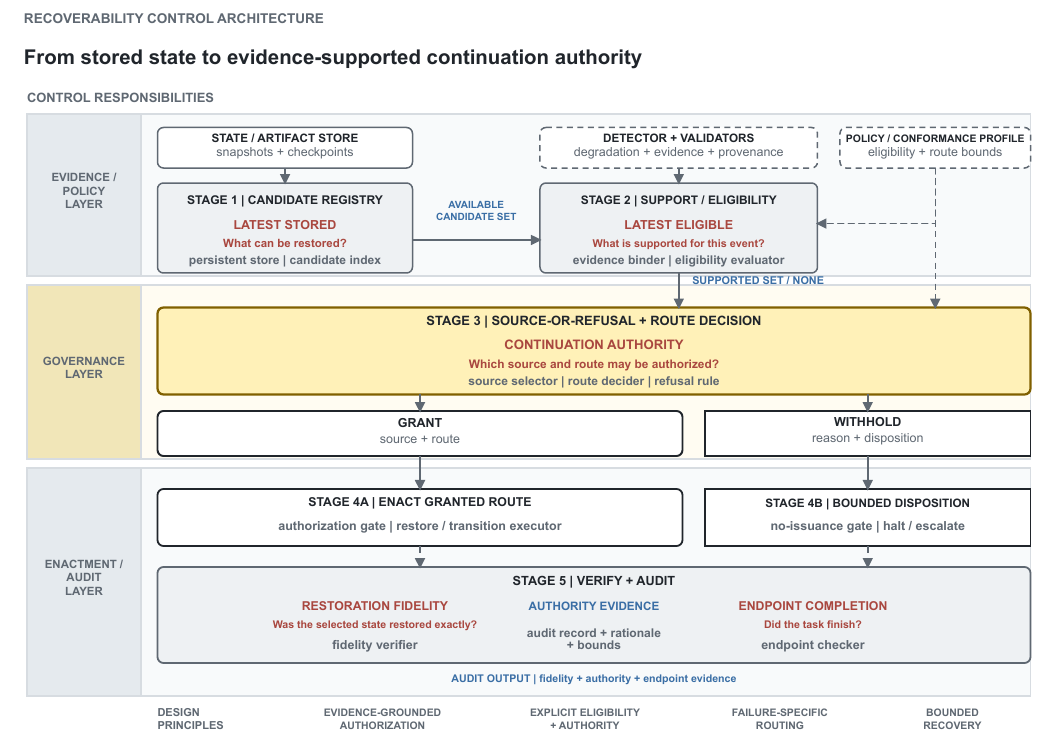}
  \caption{RPR reference control architecture. Three layers connect candidate registration and eligibility assessment to a recovery decision, its execution or withholding, and subsequent verification. Audit preserves separate evidence about restoration fidelity, adherence to the decision, and the final task outcome.}
  \label{fig:2:recoverability_control_plane_architecture}
\end{figure}
\textbf{Stage 1: register available candidates.} In the Evidence/\hspace{0pt}Policy layer, a store adapter exposes snapshots, checkpoints, or versioned artifacts together with the information needed to reconstruct or use them. The candidate index records their identities and references, producing the available set $B_t$. The \emph{latest stored} query $s_t$ returns its newest member. Registration establishes availability; a candidate can remain stored and inspectable even when it is unsuitable for continuation.

\textbf{Stage 2: assess support for the event.} The evidence binder associates the degradation event with candidate versions, validation results, provenance, and the governing policy. The eligibility evaluator uses these inputs to identify the supported subset of $B_t$, which may be empty. The \emph{latest eligible} query $\ell_t$ returns the newest member of this subset. A previous validation is relevant only within its scope and while it remains applicable. The stage supplies candidate-\hspace{0pt}level support, not permission to execute a particular action.

\textbf{Stage 3: decide the source and route, or withhold.} The Governance layer combines a source selector, route decider, and refusal rule. It consumes the supported candidates together with event evidence and route constraints, and emits one policy-\hspace{0pt}valid decision: \code{GRANT(source, route)} or \code{WITHHOLD(reason, disposition)}. A grant permits the named source-\hspace{0pt}and-\hspace{0pt}action combination. A withhold issues no continuation capability and specifies a halt or escalation disposition. The newest supported candidate need not be selected, and even a supported candidate may have no valid continuing route. The policy also determines when continuation is required, so unconditional withholding would not satisfy the contract.

\textbf{Stage 4: enact the decision.} In the Enactment/\hspace{0pt}Audit layer, the two outcomes follow different paths. Stage 4A passes a grant through an authorization gate to the restore or transition executor. Stage 4B passes withholding through a no-\hspace{0pt}issuance gate to the prescribed halt or escalation. The gate is a control responsibility that an implementation must realize: execution through the governed path must respect the permitted source, action, and bounds. A refusal record alone does not establish that this happened.

\textbf{Stage 5: verify and retain evidence.} A fidelity verifier checks whether restoration reconstructed the selected state. Authority evidence records whether execution and disposition matched the decision and its bounds, while an endpoint checker evaluates the final task invariant. The audit record connects these observations with the event, policy, and reason for the decision. It must permit the source and route to be checked as well as their execution; faithfully carrying out an invalid decision is still a recovery error.

The architecture thus distinguishes five questions: latest stored, latest eligible, continuation authority, restoration fidelity, and endpoint completion. The first two are queries about available and supported states; authority is a decision; fidelity and completion are observations about execution. Their connection is the design's central requirement: evidence constrains the decision, the decision constrains execution, and retained evidence makes that relation checkable.

\subsection{State, Candidates, and Evidence}
The architecture operates on objects with distinct roles. Operational state and artifacts describe work; a checkpoint identifies a possible recovery point; traces and verifier outputs supply evidence about what happened and what was checked. Table 2 summarizes the reference object model. These are logical roles rather than a requirement to implement one class per row.

\begin{table}[htbp]
  \centering
  \small
  \renewcommand{\arraystretch}{1.12}
  \caption{Objects and their roles in the recoverability design.}
  \label{tab:2}
  \begin{tabularx}{\linewidth}{@{}>{\hsize=0.480\hsize\linewidth=\hsize\raggedright\arraybackslash}X>{\hsize=1.520\hsize\linewidth=\hsize\raggedright\arraybackslash}X@{}}
  \toprule
\textbf{Object} & \textbf{Architectural role} \\
\midrule
\code{Operational\hspace{0pt}State} & Current goal, cursor, plan, and unresolved work needed to interpret execution. \\
\code{Artifact} & Versioned task product, such as a document, patch, or table, whose content can be inspected and validated. \\
\code{Checkpoint} & Stored state reference and continuation context forming a candidate recovery point. \\
\code{Memory\hspace{0pt}Claim} & Retained fact, reflection, or skill whose applicability must be reconciled with current evidence. \\
\code{Trace} & Record of actions, observations, and effects supporting provenance and diagnosis. \\
\code{Verifier\hspace{0pt}Output} & Result of a scoped test, schema check, policy check, or approval tied to the version it assessed. \\
\code{Evidence\hspace{0pt}Package} & Event-\hspace{0pt}specific collection of candidate, validation, provenance, and route evidence for a recovery decision. \\
\code{Recovery\hspace{0pt}Route} & Corrective action or withholding disposition, interpreted under its preconditions and effects. \\
\code{Handoff\hspace{0pt}Package} & Proposed transfer of established continuation context to another executor. \\
  \bottomrule
  \end{tabularx}
\end{table}

An \textbf{EvidencePackage} organizes four questions: what happened, which evidence is relevant, which candidate states remain supported, and which routes are available. It may include the failure event, recent history, artifact identities and versions, validation records, provenance, candidate routes, and a reason why continuing from the current state is unsupported. It collects inputs to the decision; neither the package nor an individual validation result grants continuation authority by its existence. Section 5 distinguishes evidence recorded by the implementations from evidence actually consumed in their decisions.

A \textbf{HandoffPackage} serves a later purpose. Another agent, session, or human would need the established recovery point, relevant artifacts, intended next action, and unresolved questions. A detailed diagnostic record need not make useful handoff context, and a concise handoff does not prove that its contents remain applicable. The recipient would still need to respect the conditions for continuing. This transfer is a design extension whose implementation status is stated in Section 5.

Versioned artifacts make evidence refer to specific work even after conversational context is lost. We call continuity based on a supported artifact version \textbf{artifact-\hspace{0pt}anchored continuity}. The artifact fixes the reference; validation and policy establish whether it can be reused. A newer edit, a superseded approval, or an expired check can change that judgment without erasing the saved version. With dependent artifacts, support would also need to account for their relationships: retaining one file while rolling back another may break a dependency, and memory may refer to content that no longer exists. Cross-\hspace{0pt}artifact reconciliation and partial rollback remain extensions beyond the evaluated implementations.

\subsection{Candidate Lifecycle and Recovery Decisions}
During ordinary execution, the runtime records action attempts and their resulting artifacts or effects. Storage makes candidate states available; checks provide evidence tied to those states. An implementation may record eligibility when a checkpoint is created, maintain it separately, or assess it when recovery is requested. In each case, persistence alone must not make an unverified result a permissible recovery point.

For implementations that promote candidates after validation, promotion has an ordering invariant: the output of a failed step must not advance the eligible boundary before validation and failure inspection finish. A rejected version may remain stored for inspection, while promptly recording supported progress allows later recovery to retain it. In the document example, saving exploratory B does not replace A's approval record, and checking B does not change which content A's earlier validation covered. An eligibility record made at creation is evidence for a later decision, not a substitute for considering changes in its applicability.

After degradation, the event is bound to the candidate set and the evidence relevant to reuse. If A remains supported while B is excluded, the decision procedure considers valid routes from A. An allowed retry can yield a grant from A. If the call may already have changed external state, the same document approval may be insufficient to permit retry; the policy can require further verification or withholding. If all available evidence has expired and no other supported pair exists, the runtime follows a valid non-\hspace{0pt}continuation disposition.

Source and route can be chosen jointly or reconsidered as constraints are checked. The architecture does not prescribe a rule table, planner, or learned router. It requires the chosen combination or withholding disposition to belong to the allowed decision set in Section 3, with its reason and bounds recorded. A change in failure evidence can therefore change what happens next without changing the storage mechanism. Section 5 explains the narrower decision procedures used in the implemented instances.

\subsection{Enactment, Verification, and Bounded Recovery}
A grant must identify the source and route that the executor may use. The execution path must preserve that association through restoration or transition; loading a different snapshot or performing an unpermitted action would not satisfy the decision. Withholding must leave no automatic continuing action available through that path and must lead to the prescribed disposition. These are requirements for an implementation within its declared scope, not security guarantees supplied by the diagram.

Verification then examines both the decision and its enactment. The retained record must identify the policy and evidence against which the source and route were assessed, and show what the executor actually did. Comparing only restored bytes checks fidelity, not the validity of the selected source. Comparing only the final artifact can miss an earlier recovery violation that later repair concealed. Likewise, agreement among execution records is insufficient if the expected policy comes from the same component whose behavior is being checked. The basis for accepting conformance must remain distinguishable from that component's own report.

Accounting should survive restoration of task state. The event ledger records attempts and effects already incurred, so rolling back the task does not erase them from the recovery history or reset the account of work performed. This record supports audit and cost measurement; it does not select a source or grant permission. Its integrity and trust assumptions must be established by the particular implementation.

Finally, recovery attempts, total execution, retained artifacts, and repeated routes need explicit limits and a disposition when a limit is reached. A loop that repeatedly restores a valid source can still fail to make progress. Enforcing bounds and recording their outcome therefore belong to recovery execution itself. The following runtime instances implement different parts of this reference design; their evaluated responsibilities are stated separately.

\section{Runtime Instantiations}
Three runtime instances implement distinct parts of the RPR reference design. \textbf{RPR-\hspace{0pt}Grant} selects and restores a checkpoint using recorded validation and fixed recovery rules. \textbf{RPR-\hspace{0pt}File} applies the same source-\hspace{0pt}selection rule to model-\hspace{0pt}generated documents and actual file writes. \textbf{RPR-\hspace{0pt}Authority} examines a later question: whether permission is justified at the failure event and whether execution follows it. Table 3 separates their implemented scope so that the reference architecture is not mistaken for a feature list shared by all three. The file executor is not connected to the authority apparatus in these experiments; their evidence is complementary rather than an integrated end-\hspace{0pt}to-\hspace{0pt}end validation.

\begin{table}[htbp]
  \centering
  \small
  \renewcommand{\arraystretch}{1.12}
  \caption{Implemented responsibilities and scope of the runtime instances.}
  \label{tab:3}
  \begin{tabularx}{\linewidth}{@{}>{\hsize=0.480\hsize\linewidth=\hsize\raggedright\arraybackslash}X>{\hsize=1.170\hsize\linewidth=\hsize\raggedright\arraybackslash}X>{\hsize=1.350\hsize\linewidth=\hsize\raggedright\arraybackslash}X@{}}
  \toprule
\textbf{Instance} & \textbf{State and control mechanism} & \textbf{Scope of the evaluated behavior} \\
\midrule
RPR-\hspace{0pt}Grant & Single-\hspace{0pt}executor task state, deep checkpoints, recorded eligibility, fixed recovery routes, and an external action ledger & Creation-\hspace{0pt}time contract eligibility and scoped restoration; all recovering paths have an eligible source. \\
RPR-\hspace{0pt}File & Document snapshots, a deterministic source rule, actual file restoration, and file-\hspace{0pt}tool records & The same source contract with sampled content and repairs; no general task-\hspace{0pt}state restoration or event-\hspace{0pt}time authority assessment. \\
RPR-\hspace{0pt}Authority & Supplied event-\hspace{0pt}time evidence and policy, a task-\hspace{0pt}side controller, an isolated effect authority, and retained verification evidence & Both grants and withholds, enactment, and bounds within a declared profile and trust model. \\
  \bottomrule
  \end{tabularx}
\end{table}

\subsection{RPR-\hspace{0pt}Grant: Selecting a Checkpoint Recorded as Eligible}
RPR-\hspace{0pt}Grant keeps the task cursor, history, artifact versions, validation results, checkpoint pointers, and recovery records. Its two task families edit one document or transform one structured table. A new artifact is initially ineligible for use as a recovery source. It can be promoted only when task validation and continuation policy admit it. A \textbf{working-\hspace{0pt}only} checkpoint saves the state without that promotion, including cases where structural validation passes but another continuation rule rejects the version.

The document validator checks a heading, introduction content, and scenario-\hspace{0pt}declared invariants such as heading preservation or a required transition. The table validator checks required columns, a positive row count, and declared invariants such as row-\hspace{0pt}count preservation or normalized column order. These are task-\hspace{0pt}specific structural and content checks, not general semantic judgments. Their results inform a separate checkpoint policy: \code{validator\_\hspace{0pt}gated} permits promotion after a promotable validation, whereas \code{working\_\hspace{0pt}only} stores a checkpoint without advancing the eligible pointer, including when validation passes. The runtime records whether exclusion follows from validation or continuation policy.

The distinction is recorded when a checkpoint is created. Let $K$ be the task contract fixed before execution, and let $E_K(b,X_{\operatorname{create}(b)})$ report whether candidate $b$ met its conditions at that time. The latest recorded eligible checkpoint is

\begin{equation*}
e_t^K=\max_{\prec}\{b\in B_t:E_K(b,X_{\operatorname{create}(b)})=1\},
\end{equation*}

with $e_t^K=\varnothing$ if none was admitted. The runtime stores this pointer separately from the latest saved checkpoint $s_t$. It does not recheck event-\hspace{0pt}time eligibility $\ell_t$; the two would coincide only if the earlier evidence still applied. Code names such as \code{trusted\_\hspace{0pt}checkpoint\_\hspace{0pt}id}, \code{last\_\hspace{0pt}trusted\_\hspace{0pt}checkpoint}, and \code{resume\_\hspace{0pt}retry\_\hspace{0pt}from\_\hspace{0pt}trusted\_\hspace{0pt}checkpoint} are historical identifiers for this recorded, contract-\hspace{0pt}relative status. They do not mean that the runtime infers objective trust.

To reuse a selected checkpoint, the runtime restores a deep snapshot of artifact versions, history, validation, progress, metadata, and continuation pointers. It replaces the corresponding working state, removes later checkpoints, resets the cursor, and clears transient failure objects. Comparing expected and restored digests checks that this was an actual state restoration, rather than a cursor change alone. Restoring a working-\hspace{0pt}only checkpoint does not promote it to eligible status.

Three injected failures exercise the loop: \code{execution\_\hspace{0pt}error}, \code{action\_\hspace{0pt}repetition}, and \code{semantic\_\hspace{0pt}stagnation}. The first occurs after an action begins but before it produces an artifact. The other two alter observable repetition or progress state, which a watchdog checks before promoting that step's artifact. The resulting EvidencePackage records the failure, eligible checkpoint, and recent history, artifacts, and validation. The router uses only the failure family and recorded eligible checkpoint; retaining other fields does not establish that they improve its decisions.

The four execution modes use the same task, failure injection, validation, and checkpoint machinery. Table 4 summarizes their different responses after a failure.

\begin{table}[htbp]
  \centering
  \small
  \renewcommand{\arraystretch}{1.12}
  \caption{Execution modes and post-\hspace{0pt}failure source policies.}
  \label{tab:4}
  \begin{tabularx}{\linewidth}{@{}>{\hsize=0.440\hsize\linewidth=\hsize\raggedright\arraybackslash}X>{\hsize=1.560\hsize\linewidth=\hsize\raggedright\arraybackslash}X@{}}
  \toprule
\textbf{Mode} & \textbf{Response and continuation source} \\
\midrule
Fail-\hspace{0pt}fast & Terminate without continuation. \\
Blind retry & Restart at step 1 and clear the remaining scripted injection. \\
Checkpoint-\hspace{0pt}only & Restore the latest stored checkpoint with a generic transition. \\
Recoverability & Restore the latest recorded contract-\hspace{0pt}eligible checkpoint with a fixed failure-\hspace{0pt}specific transition. \\
  \bottomrule
  \end{tabularx}
\end{table}

The recoverability mode maps these failures to restore-\hspace{0pt}then-\hspace{0pt}retry, restore-\hspace{0pt}then-\hspace{0pt}replan, and rollback-\hspace{0pt}restore-\hspace{0pt}then-\hspace{0pt}replan. The scenario supplies the repair content; the runtime chooses the source and fixed transition. Because source and route rules change together between the restoring modes, their effects are not fully isolated. All evaluated recovering paths have an eligible source. The implementation also contains a \code{halt} label, but the presence of that label is not evidence that withholding was exercised.

An event ledger counts work that recovery must not erase from the record: action attempts, artifact generation, validation, checkpoint operations, decisions, restarts, restorations, and bound violations. It is kept outside restorable task state, so a rollback cannot remove already incurred attempts. The ledger measures and audits execution; it neither selects eligible checkpoints nor authorizes actions. Guardrails limit recovery to three attempts and executed steps and materialized artifacts to $4q$, where $q$ is the maximum of the scenario step limit, configured sequence length, and one.

RPR-\hspace{0pt}Grant also defines a HandoffPackage schema and state serialization helpers, but the evaluated loop neither constructs nor consumes a handoff and does not exercise persistent-\hspace{0pt}session reload. Memory-\hspace{0pt}conflict resolution is not implemented. These definitions provide extension points; their presence does not establish cross-\hspace{0pt}session recovery or full conformance to the reference design.

\subsection{RPR-\hspace{0pt}File: Model-\hspace{0pt}Generated Documents and File Restoration}
RPR-\hspace{0pt}File tests the same distinction when the document content comes from a model and restoration changes an actual file. It snapshots approved and working-\hspace{0pt}only versions, writes the selected snapshot back to disk, and verifies its SHA-256 digest before a separate model call proposes a repair. The record therefore distinguishes the chosen starting point, the restored file, and the repaired result. Model calls supply content; they do not decide eligibility or select a route. RPR-\hspace{0pt}Grant and RPR-\hspace{0pt}File are frozen separately as artifacts v03 and v04. Their scope is recorded eligibility and restoration, without general evidence assessment, cross-\hspace{0pt}session handoff, or full-\hspace{0pt}contract conformance.

\subsection{RPR-\hspace{0pt}Authority: Checking Permission at the Failure Event}
RPR-\hspace{0pt}Authority considers conditions at the failure event rather than relying on an earlier promotion record. Its supplied facts and rules can require continuation, make every source ineligible, or leave an eligible source with no allowed action. A controller decides in the task process. A separate oracle interprets the same frozen specification without importing the controller's rules and supplies an independent reference for $\mathcal{D}_t$. This oracle scores decisions; it grants no runtime permission. Twenty-\hspace{0pt}six frozen contracts specify the input, decision, protocol, evidence, and record formats. Expected values remain protected until the manifest is verified.

To make permission affect execution, each run uses a fresh isolated authority process. That process owns the state being changed, the effect ledger, recovery bounds, and the rules mapping routes to effect handlers, called sinks. It issues one-\hspace{0pt}use capabilities through which the task process can request effects over a line-\hspace{0pt}framed JSON channel. Canonical spawn arguments fix the session configuration, and later effect messages cannot choose a different run identity. Malformed or noncanonical frames are sealed before application; a capability is consumed atomically before dispatch. The process records a hash-\hspace{0pt}chained transcript, final state, and seal for independent checking.

The check also needs an expected policy that the task-\hspace{0pt}side client cannot redefine. The trusted caller constructs the canonical configuration and retains an independent deep copy before creating any client object. Verification compares the configuration in the authority-\hspace{0pt}written evidence with this earlier snapshot. A replaced client may cause a different configuration to be executed, but it cannot make that configuration agree with the independently retained reference merely by reporting it back. Section 6 examines this distinction experimentally.

The frozen apparatus uses the following threat boundary:

\begin{quote}
\textbf{Threat model:} apparatus-\hspace{0pt}channel adversary. Probes run in the task process and may perform arbitrary operations on apparatus-\hspace{0pt}exposed entry points and channels, including rewriting or replacing client-\hspace{0pt}stub objects and methods, \code{vars()}, \code{object.\_\hspace{0pt}\_\hspace{0pt}setattr\_\hspace{0pt}\_\hspace{0pt}}, class replacement, and any other ordinary Python operation, as well as sending arbitrary byte sequences over the channel (forged, replayed, out-\hspace{0pt}of-\hspace{0pt}order, or unauthorized messages).
\textbf{Claim:} under this adversary, no new effect occurs outside the authority boundary. This fact is established solely by independent recomputation from retained bytes written by the authority process itself (a hash-\hspace{0pt}chained transcript, final state, and seal, all three bound to the candidate tree); no task-\hspace{0pt}process self-\hspace{0pt}report is trusted.
\textbf{Explicitly out of scope:} integrity of the spawn mechanism (the client module and its subprocess-\hspace{0pt}creation path, on the same trust base and standard as the oracle subprocess); OS-\hspace{0pt}level process injection or debugging interfaces; and post hoc tampering with retained bytes (covered separately by the RUN\_MANIFEST, tree identity, Git, and governance ledger).
\end{quote}

The statement above concerns effects mediated and recorded by the authority process. It does not imply that every effect caused through a replaced client is prevented. Such a client can launch a real authority process under an expanded configuration; comparison with the caller's retained configuration then determines whether the resulting execution can be accepted as conforming. Integrity of the trusted caller and spawn mechanism is assumed. Replacing the caller-\hspace{0pt}owned configuration builder is outside scope, and the controller's recorded projection of configuration is not claimed to detect every possible change in sink authorization.

\section{Controlled Evaluation}
\subsection{Research Questions and Study Roles}
The evaluation follows the practical recovery questions in order: choose a suitable checkpoint, retain the work it contains, repeat the check with actual files, and test when execution must be allowed or stopped. Table 5 gives the role and scope of each study using the runtimes introduced in Section 5.

\begin{table}[htbp]
  \centering
  \small
  \renewcommand{\arraystretch}{1.12}
  \caption{Research questions and controlled study scope.}
  \label{tab:5}
  \begin{tabularx}{\linewidth}{@{}>{\hsize=0.270\hsize\linewidth=\hsize\raggedright\arraybackslash}X>{\hsize=1.500\hsize\linewidth=\hsize\raggedright\arraybackslash}X>{\hsize=1.230\hsize\linewidth=\hsize\raggedright\arraybackslash}X@{}}
  \toprule
\textbf{Study} & \textbf{Research question} & \textbf{Runtime and controlled scope} \\
\midrule
A & When the newest checkpoint is ineligible, does recovery choose the allowed one, and would checking the final result reveal a wrong choice? & RPR-\hspace{0pt}Grant; 16 scenarios, 64 runs \\
B & When both restoring modes choose the same checkpoint, how much validated work is retained and how many actions are repeated? & RPR-\hspace{0pt}Grant; 18 scenarios, 72 runs \\
C & Does the same source-\hspace{0pt}choice distinction hold for shared model-\hspace{0pt}generated candidates and actual file restoration? & RPR-\hspace{0pt}File; 20 retained candidate pairs, 40 runs \\
D & When current conditions require continuation or withholding, does execution follow that decision, and can its record be checked independently? & RPR-\hspace{0pt}Authority; 18 matched cases, qualification and a 36-\hspace{0pt}run main matrix \\
  \bottomrule
  \end{tabularx}
\end{table}

Studies A and B use scripted content to isolate runtime behavior. Study C adds model-\hspace{0pt}generated documents and repairs while keeping source rules fixed. All three use the eligibility recorded at checkpoint creation and have an eligible source in their recovering paths. Study D separately changes the supplied conditions at the failure event. These constructed studies address different requirements, so we report their exact counts and controls separately rather than pooling them into a benchmark score.

The scoring follows each declared profile. A/\hspace{0pt}C require a particular checkpoint, so source accuracy compares the chosen identifier with that reference. The general contract in Section 3 permits any source-\hspace{0pt}and-\hspace{0pt}route pair in the allowed set; choosing the newest eligible checkpoint is not a universal conformance requirement. Restoration and final-\hspace{0pt}invariant checks then assess different observations, without retrospectively authorizing the starting choice.

\begin{figure}[htbp]
  \centering
  \includegraphics[width=\linewidth]{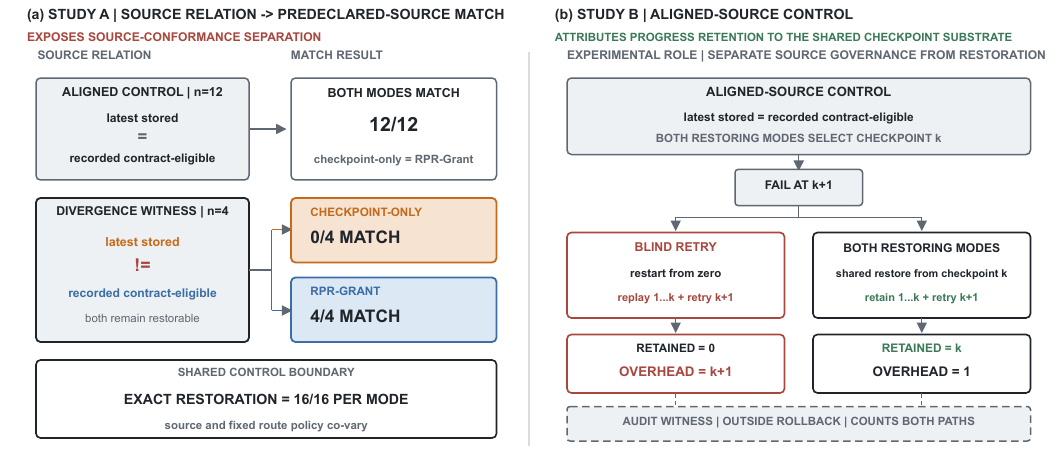}
  \caption{Choosing a recovery point and preserving its progress. (a) The modes choose differently when the newest saved checkpoint is ineligible. (b) When both choose the same checkpoint, both preserve its validated prefix. Tables 6 and 7 separate source-\hspace{0pt}choice results from action accounting.}
  \label{fig:3:boundary_selection_and_cost}
\end{figure}
\subsection{Study A: Choosing an Eligible Recovery Point}
\textbf{Design.} Study A makes the choice of recovery point visible while keeping restoration shared. Twelve original scenarios combine two task families, three failure families, and recovery from an initial or later eligible checkpoint. In these cases the latest stored checkpoint is also the latest recorded eligible one. Four additional challenges deliberately separate them: each stores eligible A, then newer working-\hspace{0pt}only B, before a later failure. The challenges cross the two tasks with two reasons for excluding B: failed validation, or policy rejection or supersession despite structural validation. Each scenario fixes the required source and acceptable route before execution.

Each scenario runs in the four modes described in Section 5.1, for 64 runs. The two restoring modes share checkpoint creation, snapshot format, validators, task execution, restoration, and guardrails. Checkpoint-\hspace{0pt}only selects $s_t$ and its generic route; recoverability selects $e_t^K$ and a fixed failure-\hspace{0pt}specific route. Source and route policy therefore co-\hspace{0pt}vary. The selected source identifier is directly observable, but the comparison does not independently identify the causal contribution of routing.

\textbf{Measurements.} The \textbf{contract-\hspace{0pt}gold source} is the checkpoint required by the predeclared task contract. Source selection checks the chosen identifier against it. Restoration fidelity checks the selected snapshot against its digest manifest, and final-\hspace{0pt}invariant satisfaction checks the resulting artifact. The run-\hspace{0pt}completed flag reports termination state and is kept separate from that final check. Candidate availability and route records are also retained. The legacy mode-\hspace{0pt}gated \code{trusted\_\hspace{0pt}anchor\_\hspace{0pt}accuracy} is not used for the neutral source comparison.

\begin{table}[htbp]
  \centering
  \small
  \renewcommand{\arraystretch}{1.12}
  \caption{Source selection, restoration, and endpoints in Study A. The table reports the two restoring modes; fail-\hspace{0pt}fast and blind retry are included in the 64-\hspace{0pt}run matrix but do not select and restore checkpoints.}
  \label{tab:6}
  \begin{tabularx}{\linewidth}{@{}>{\hsize=1.380\hsize\linewidth=\hsize\raggedright\arraybackslash}X>{\hsize=1.020\hsize\linewidth=\hsize\raggedright\arraybackslash}X>{\hsize=0.900\hsize\linewidth=\hsize\raggedleft\arraybackslash}X>{\hsize=0.900\hsize\linewidth=\hsize\raggedleft\arraybackslash}X>{\hsize=0.900\hsize\linewidth=\hsize\raggedleft\arraybackslash}X>{\hsize=0.900\hsize\linewidth=\hsize\raggedleft\arraybackslash}X@{}}
  \toprule
\textbf{Scenario group} & \textbf{System} & \textbf{Contract-\hspace{0pt}gold source selected} & \textbf{Exact restoration} & \textbf{Run completed flag} & \textbf{Final invariants} \\
\midrule
Original scenarios ($n=12$) & Checkpoint-\hspace{0pt}only & 12/12 & 12/12 & 12/12 & 10/12 \\
Original scenarios ($n=12$) & Recoverability & 12/12 & 12/12 & 12/12 & 12/12 \\
Eligibility challenges ($n=4$) & Checkpoint-\hspace{0pt}only & 0/4 & 4/4 & 4/4 & 4/4 \\
Eligibility challenges ($n=4$) & Recoverability & 4/4 & 4/4 & 4/4 & 4/4 \\
  \bottomrule
  \end{tabularx}
\end{table}

\textbf{Results.} When stored and recorded-\hspace{0pt}eligible sources align, both restoring modes select the required source in all 12 cases. When they diverge, checkpoint-\hspace{0pt}only follows recency and selects B in all four challenges, whereas recoverability selects A. Both modes restore every selected snapshot exactly. The source contrast is therefore not explained by unequal restoration fidelity.

Both modes also complete and satisfy the final invariant in all four challenges. Thus checking restored bytes or the final document would miss checkpoint-\hspace{0pt}only's selection of an excluded source. The source-\hspace{0pt}score difference follows from the constructed challenge and declared policies; it is not evidence of learned selection or a higher final success rate. Its value is a controlled counterexample to treating fidelity or task success as sufficient evidence of a correct recovery choice. The experiment checks the creation-\hspace{0pt}time contract; it does not independently reassess whether earlier evidence remains valid at failure.

The two checkpoint-\hspace{0pt}only invariant misses occur in the original, aligned-\hspace{0pt}source scenarios. Their source is correct, but the generic route does not select the scenario's route-\hspace{0pt}specific repair payload. These observations motivate treating source and route separately; they are not an independent route ablation or evidence that the runtime generates better repairs.

\subsection{Study B: Preserving Validated Progress}
\textbf{Design.} Study B asks how much work must be repeated when the chosen checkpoint is already agreed upon. It varies the validated prefix before failures at steps 2, 4, and 8, across the two task families and three failure families, giving 18 scenarios and 72 runs. Every successful pre-\hspace{0pt}failure step validates and creates an eligible checkpoint. Thus $s_t=e_t^K$, and both restoring modes choose checkpoint $k\in\{1,3,7\}$ just before the failed action at $k+1$.

Checkpoint spacing is one action. Blind retry restarts from step 1 and clears the single scripted failure, so it can eventually complete. All continuing modes have valid post-\hspace{0pt}recovery task steps. Holding the checkpoint choice constant lets the study attribute retained work to restoration, rather than to a different source policy. The manipulated quantity is prefix length, not general task difficulty.

\textbf{Measurements.} A common ledger records incurred action attempts outside restorable state, including the failing attempt. We compute

\begin{equation*}
\begin{aligned}
\mathrm{ReexecutionOverhead}={}&\mathrm{TotalActionAttempts}\\
&-\mathrm{UsefulFinalTrajectoryActions},
\end{aligned}
\end{equation*}

where useful final-\hspace{0pt}trajectory actions are distinct positive step indices represented by contract-\hspace{0pt}eligible artifacts retained in the final working state. We also count validated actions in the selected snapshot at first recovery. The ledger supports measurement and audit; it does not authorize either source.

\begin{table}[htbp]
  \centering
  \small
  \renewcommand{\arraystretch}{1.12}
  \caption{Action accounting in Study B. Each row averages the two task families and three failure families. The frozen shallow/\hspace{0pt}medium/\hspace{0pt}deep labels denote configured step counts. Fail-\hspace{0pt}fast does not continue and is omitted from this table.}
  \label{tab:7}
  \begin{tabularx}{\linewidth}{@{}>{\hsize=1.080\hsize\linewidth=\hsize\raggedright\arraybackslash}X>{\hsize=0.960\hsize\linewidth=\hsize\raggedright\arraybackslash}X>{\hsize=0.780\hsize\linewidth=\hsize\raggedleft\arraybackslash}X>{\hsize=1.020\hsize\linewidth=\hsize\raggedleft\arraybackslash}X>{\hsize=1.020\hsize\linewidth=\hsize\raggedleft\arraybackslash}X>{\hsize=1.140\hsize\linewidth=\hsize\raggedleft\arraybackslash}X@{}}
  \toprule
\textbf{Benchmark position label} & \textbf{System} & \textbf{Total attempts} & \textbf{Useful final actions} & \textbf{Re-\hspace{0pt}execution overhead} & \textbf{Validated actions retained at recovery} \\
\midrule
Shallow & Blind retry & 4.0 & 2.0 & 2.0 & 0.0 \\
Shallow & Checkpoint-\hspace{0pt}only & 3.0 & 2.0 & 1.0 & 1.0 \\
Shallow & Recoverability & 3.0 & 2.0 & 1.0 & 1.0 \\
Medium & Blind retry & 8.0 & 4.0 & 4.0 & 0.0 \\
Medium & Checkpoint-\hspace{0pt}only & 5.0 & 4.0 & 1.0 & 3.0 \\
Medium & Recoverability & 5.0 & 4.0 & 1.0 & 3.0 \\
Deep & Blind retry & 16.0 & 8.0 & 8.0 & 0.0 \\
Deep & Checkpoint-\hspace{0pt}only & 9.0 & 8.0 & 1.0 & 7.0 \\
Deep & Recoverability & 9.0 & 8.0 & 1.0 & 7.0 \\
  \bottomrule
  \end{tabularx}
\end{table}

\textbf{Results.} Blind retry's overhead grows from 2 to 4 to 8 attempts as it repeats the growing prefix. Both restoring modes retain 1, 3, and 7 validated actions and incur one extra attempt. This pattern holds in every task-\hspace{0pt}by-\hspace{0pt}failure-\hspace{0pt}by-\hspace{0pt}position cell, rather than arising from compensating subgroup averages. All 36 restoring runs have exact fidelity, all continuing modes satisfy final invariants, and none of the 72 runs triggers a guardrail.

The equal restoring-\hspace{0pt}mode results explain the value of checkpoint preservation: both reuse the validated prefix instead of repeating it. This benefit belongs to their shared restoration mechanism. Recoverability adds a decision about when such reuse is allowed, as Study A tests, rather than a separate saving in this aligned-\hspace{0pt}source case. The measurements are action counts under unit spacing, not latency, token cost, checkpoint I/\hspace{0pt}O, or production efficiency.

\subsection{Study C: Shared Candidates and Actual File Restoration}
\textbf{Design.} Study C asks whether the same source-\hspace{0pt}choice distinction remains visible when the runtime works with model-\hspace{0pt}generated documents on disk. Four short tasks protect a phrase, a Markdown heading, a citation token, or a numeric recovery limit, with five paired repetitions per task. A model rewrites approved A under an instruction to omit the invariant. Only a validator-\hspace{0pt}rejected B is retained, with one additional generation attempt allowed if needed to create that challenge. Unretained attempts are not serialized.

Each retained candidate's exact digest is shared across two independent mode workspaces. Both modes write A and B, snapshot their contents, and encounter a formatter that writes the file before a simulated timeout leaves tool status ambiguous. Checkpoint-\hspace{0pt}only restores B; recoverability restores the contract-\hspace{0pt}eligible A. Restoration writes the selected snapshot to disk and checks its SHA-256 digest. Each mode then makes an independent model call to repair the restored document and validates the resulting file. The repair calls use different restored inputs; source selection itself remains deterministic.

\begin{figure}[htbp]
  \centering
  \includegraphics[width=\linewidth]{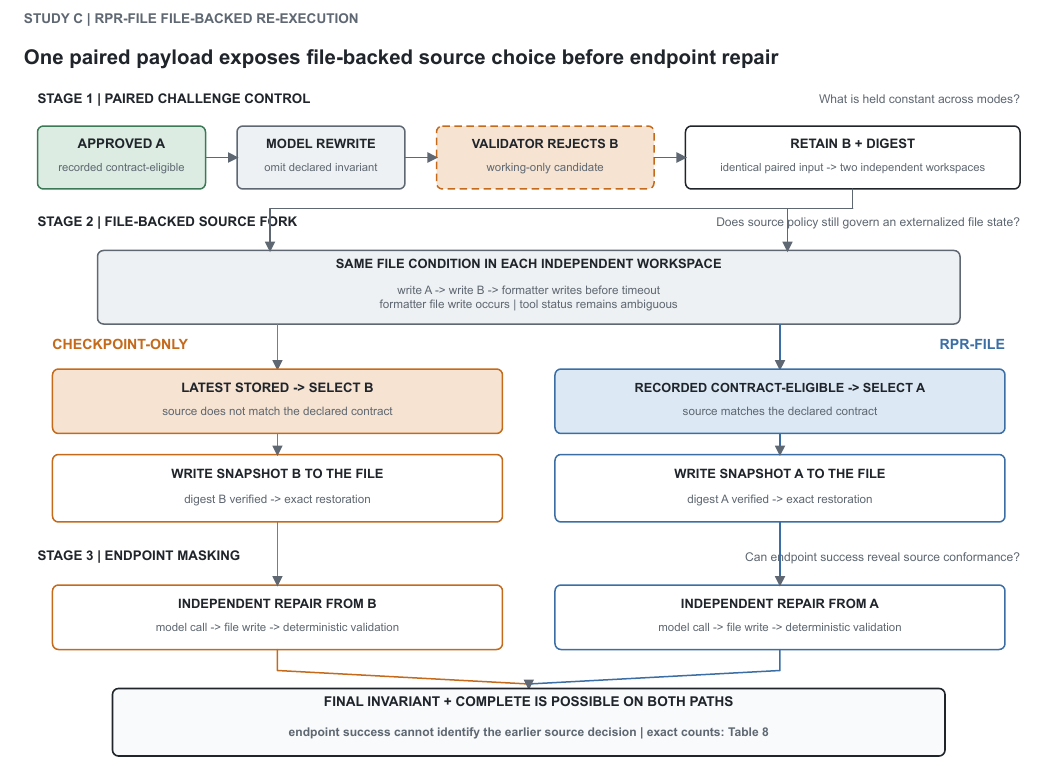}
  \caption{Paired file-\hspace{0pt}restoration protocol. Each mode receives the same retained candidate B and the same write-\hspace{0pt}then-\hspace{0pt}timeout pattern. Their source policies select different snapshots, which are restored and checked before independent repair. Successful final files do not retrospectively establish that both sources were allowed.}
  \label{fig:4:study_c_paired_file_transfer}
\end{figure}
\begin{table}[htbp]
  \centering
  \small
  \renewcommand{\arraystretch}{1.12}
  \caption{Paired source selection and file outcomes in Study C. Each of the 20 retained challenge candidates is shared across modes.}
  \label{tab:8}
  \begin{tabularx}{\linewidth}{@{}>{\hsize=1.320\hsize\linewidth=\hsize\raggedright\arraybackslash}X>{\hsize=0.480\hsize\linewidth=\hsize\raggedleft\arraybackslash}X>{\hsize=1.080\hsize\linewidth=\hsize\raggedleft\arraybackslash}X>{\hsize=1.020\hsize\linewidth=\hsize\raggedleft\arraybackslash}X>{\hsize=1.020\hsize\linewidth=\hsize\raggedleft\arraybackslash}X>{\hsize=1.080\hsize\linewidth=\hsize\raggedleft\arraybackslash}X@{}}
  \toprule
\textbf{System} & \textbf{Runs} & \textbf{Contract-\hspace{0pt}gold source selected} & \textbf{Exact restoration} & \textbf{Run completed flag} & \textbf{Final invariants} \\
\midrule
Checkpoint-\hspace{0pt}only & 20 & 0/20 & 20/20 & 20/20 & 20/20 \\
Recoverability & 20 & 20/20 & 20/20 & 20/20 & 20/20 \\
  \bottomrule
  \end{tabularx}
\end{table}

\textbf{Results.} All 20 shared candidates violate their target invariant. Checkpoint-\hspace{0pt}only selects B in every pair, whereas recoverability selects A. Every selected snapshot is restored exactly, every run completes, and every repaired file satisfies its invariant. A model can therefore repair the final file successfully without revealing that the runtime started from a disallowed version. Checking the chosen recovery point separately remains useful with actual file effects and sampled repairs.

This is a within-\hspace{0pt}project transfer check of a supplied contract, not an independent replication or a test of learned source selection. The challenge candidates are selected for rejection, the validators are exact-\hspace{0pt}string checks, and the task edits one local file. The endpoint reported model ID \code{gpt-\hspace{0pt}5.6-\hspace{0pt}luna}; its upstream model and serving infrastructure were not independently verified. The retained record includes source and content hashes, restoration checks, generation records, a seven-\hspace{0pt}event file-\hspace{0pt}tool trace per run, and 40 workspaces. Appendix A gives the call settings and retention limits. Recorded latency and tokens describe this execution but do not support a controlled cost comparison.

\subsection{Study D: Enforcing and Checking Continuation Decisions}
\textbf{Design.} A previously admitted checkpoint may no longer support continuation when the failure occurs. Study D tests this question through six matched groups of three cases. Each group starts with byte-\hspace{0pt}identical fixture data; two declared pointers then control current evidence and route conditions. Case \textbf{G} retains the base conditions and requires grant. Case \textbf{W-\hspace{0pt}E} moves event time past all evidence-\hspace{0pt}validity windows, leaving no supported source and requiring withholding. Case \textbf{W-\hspace{0pt}R} keeps an eligible source but introduces an unresolved external effect that rules out every continuing route. These 18 cases distinguish an unsuitable recovery point from the absence of an allowed action, while requiring both grant and withhold behavior.

Qualification executes each case once. The main matrix executes each twice, giving 36 primary runs: 12 grant-\hspace{0pt}required and 24 withhold-\hspace{0pt}required executions. Separate semantic, bypass, and bound probes assess the apparatus and scoring path. These stages have different evidentiary roles and are not pooled as independent samples.

\begin{table}[htbp]
  \centering
  \small
  \renewcommand{\arraystretch}{1.12}
  \caption{Declared-\hspace{0pt}profile decisions and supporting validation in Study D.}
  \label{tab:9}
  \begin{tabularx}{\linewidth}{@{}>{\hsize=0.720\hsize\linewidth=\hsize\raggedright\arraybackslash}X>{\hsize=0.990\hsize\linewidth=\hsize\raggedright\arraybackslash}X>{\hsize=1.290\hsize\linewidth=\hsize\raggedright\arraybackslash}X@{}}
  \toprule
\textbf{Stage} & \textbf{Retained result} & \textbf{Evidentiary role} \\
\midrule
Seal & Pass; protected files recomputed with zero drift & Integrity check before expected values are released \\
Semantic controls (\code{sc-\hspace{0pt}30}) & 30/30 expected rejects & Scoring discrimination against controlled mutations, with independent-\hspace{0pt}process oracle output \\
Enforcement probes (\code{ep-\hspace{0pt}84}) & 84/84 blocked & Synthetic apparatus-\hspace{0pt}channel bypass probes, each with an isolated authority session \\
Recovery-\hspace{0pt}bound probes (\code{rb-\hspace{0pt}18}) & 18/18 terminated & Case-\hspace{0pt}bound primary, adversarial, protocol-\hspace{0pt}violation, and bound+1 sessions \\
Qualification (\code{qual-\hspace{0pt}18}) & 18/18 \code{ok} & One execution per matched case \\
Main matrix (\code{main-\hspace{0pt}36}) & Decisions 36/36; grants 12/12; withholds 24/24 & Two executions per case, including physical enactment and enforced dispositions \\
  \bottomrule
  \end{tabularx}
\end{table}

\textbf{Decision and execution results.} Each grant issues a one-\hspace{0pt}use capability for the allowed route, and the authority process records the resulting effect. Each withhold issues no continuing capability; unauthorized helper and token attempts remain blocked, and a second attempt terminates under the integrated bound. These observations test whether permission changes what can execute. A correct decision label without the corresponding effect or refusal would be insufficient.

Each qualification/\hspace{0pt}main record links identity, input, expected reference, decision, execution, bounds, and outcome. Independent offline verification replays the authority state machine, compares its configuration with the caller's earlier snapshot, and reconciles the run identity, bound, controller projection, route, decision, and sink authorization. The evidence comes from the process that owns the effects; task-\hspace{0pt}side success claims are not the evidence of enactment. Appendix A.3 gives the verification inventory and identifiers for the following comparison.

\textbf{Who supplies the expected policy?} A record can be internally consistent and still describe the wrong policy. The adversarial comparison changes where verification obtains its expected configuration. With a \textbf{client-\hspace{0pt}supplied reference} (round 12), that reference is copied after client construction. A replacement client expands the handlers permitted for the recovery route and causes the real authority subprocess to advance its cursor from 0 to 1. The reference changes with the client: the record remains internally consistent, and the run is accepted despite departing from the caller's intended policy.

With a \textbf{caller-\hspace{0pt}held reference} (round 13), the canonical configuration is retained before any client exists. The same attack still changes the real authority-\hspace{0pt}plane cursor, but it cannot alter this earlier reference through the client. Comparison with the authority-\hspace{0pt}written evidence detects the policy mismatch, and the in-\hspace{0pt}run check rejects conformance. The offline verifier independently recomputes the rejection. The distinguishing property is independence of the checking basis from the component under inspection; successful execution and internally coherent records cannot supply that independence.

\begin{figure}[htbp]
  \centering
  \includegraphics[width=\linewidth]{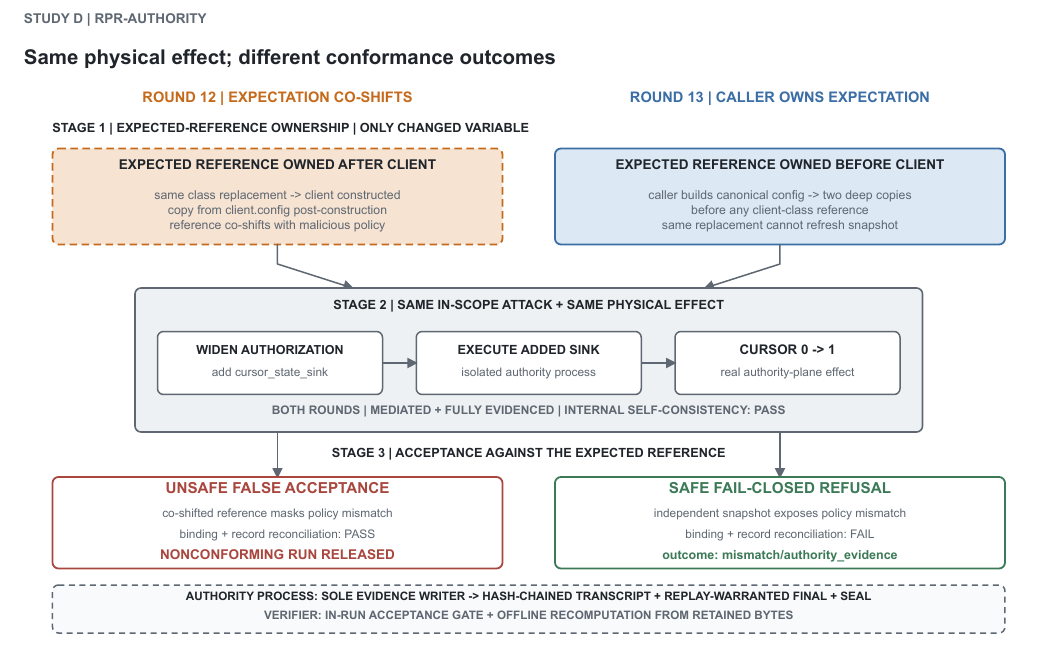}
  \caption{Independent policy evidence changes acceptance. Both rounds produce the physical effect. The caller's pre-\hspace{0pt}client reference exposes the altered policy and rejects conformance in round 13; it does not prevent or roll back the effect.}
  \label{fig:5:study_d_failure_closed_acceptance}
\end{figure}
The main matrix's \code{WITHHOLD} stops capability issuance before execution. This adversarial check instead rejects a run after its effect is recorded. Its value is that an effect and a self-\hspace{0pt}consistent record cannot make an altered policy count as the expected one when the reference is held independently. The demonstrated effect was neither prevented nor rolled back. The result depends on the Section 5.3 trust base and excludes replacement of the trusted caller or spawn mechanism.

\subsection{What the Studies Establish Together}
The studies address different requirements of the contract in Section 3.5. Table 10 brings their evidence together without treating one observation as a substitute for another. In particular, checking the restored state cannot determine whether the source was permissible, and a correct decision label cannot establish that execution obeyed it.

\begin{table}[htbp]
  \centering
  \small
  \renewcommand{\arraystretch}{1.12}
  \caption{Evidence for the four recoverability requirements and its limits.}
  \label{tab:10}
  \begin{tabularx}{\linewidth}{@{}>{\hsize=0.690\hsize\linewidth=\hsize\raggedright\arraybackslash}X>{\hsize=1.230\hsize\linewidth=\hsize\raggedright\arraybackslash}X>{\hsize=1.080\hsize\linewidth=\hsize\raggedright\arraybackslash}X@{}}
  \toprule
\textbf{Requirement} & \textbf{Evidence provided} & \textbf{Boundary of the evidence} \\
\midrule
Distinguish saved state from eligible state & A/\hspace{0pt}C separate stored from recorded-\hspace{0pt}eligible candidates; D varies event-\hspace{0pt}time evidence support. & Eligibility rules and facts are supplied; A/\hspace{0pt}C do not independently reassess support at failure. \\
Decide whether and how to continue & A/\hspace{0pt}C expose selected sources, A/\hspace{0pt}B record fixed routes, and D tests required grants and withholds. & Source and route policies co-\hspace{0pt}vary in A/\hspace{0pt}B; D tests only a declared deterministic profile. \\
Execute according to the decision & A/\hspace{0pt}C check exact restoration; D checks enacted effects, withholding, and unauthorized requests. & Restoration is scoped to task snapshots or one file; D relies on its caller/\hspace{0pt}spawn trust base. \\
Retain evidence and limit recovery & B checks common action accounting; D verifies retained authority evidence and termination under bounds. & B measures action counts, not operational cost; offline checks establish the retained record within its declared scope. \\
  \bottomrule
  \end{tabularx}
\end{table}

Study B explains why retaining a checkpoint is useful: it preserves validated work that restarting would repeat. Studies A and C explain why reuse also requires a separate decision: the newest stored version can be excluded, yet restoring it and repairing it later can still yield a successful endpoint. Study D adds event-\hspace{0pt}time permission, its enforcement, and independent checking of the policy used in execution. Together these results connect preservation of progress with explicit conditions for its reuse, while keeping restoration, decision correctness, and completion separately observable.

The studies are controlled demonstrations, so perfect scores indicate agreement with their constructed contracts. They do not establish robustness under distribution shift, independent route-\hspace{0pt}policy utility, or open-\hspace{0pt}world trust assessment. Nor do the separate implementations jointly establish a file-\hspace{0pt}agent control path with event-\hspace{0pt}time revalidation and enforcement. The reusable finding is the need to evaluate the choice and permission to continue separately from restoring bytes and finishing the task. This applies in both directions: a required withhold can leave a task unfinished while satisfying the recovery contract, and a completed task can violate it.

\section{Discussion}
\subsection{Making Reuse a Decision}
For a system designer, the main implication is to distinguish retaining work from deciding to act on it. Files, snapshots, and validation records can survive an interrupted execution, but their continued existence does not resolve whether they remain applicable. An explicit recovery decision connects the version being reused, the evidence supporting it, and the next action. This connection matters even when later repair produces an acceptable final result.

The contract also gives different components a common decision to inspect. Storage supplies available states; validators and policy supply conditions; planning supplies possible actions; the executor follows the permitted choice. This separates an error in source admission from inaccurate reconstruction, an impermissible route, or execution that departs from its grant. A more complete checkpoint improves what can be reconstructed, while a freshness check can change whether that same checkpoint may be used now. The responsibility spans components even when a more capable model supplies some of their judgments; the interface makes the relationship between those judgments and execution observable.

This interface can be implemented inside a checkpoint manager, workflow engine, or separate runtime. Its value is that the choice becomes an inspectable system output. The paper's claim does not depend on a particular module name or on excluding systems that already provide the behavior. A direct comparison should determine which requirements those systems satisfy, not assume that they lack recovery governance.

\subsection{Binding Checks to the Work They Support}
The document example illustrates a practical discipline: a validation result must refer to a particular version and scope. Passing a check on A should not silently approve later B, and retaining that result should not imply that it remains current indefinitely. A runtime can record both the artifact and the conditions under which it was accepted, then reconsider them when an event changes the basis for continuation. This matters in sustained tasks because artifacts, approvals, and external effects can outlive the execution context that produced them. It motivates the interface without implying that the present short tasks establish general long-\hspace{0pt}horizon performance.

The present grant-\hspace{0pt}path implementations use creation-\hspace{0pt}time records. They demonstrate the benefit of distinguishing those records from the latest saved state, but do not solve later evidence assessment. More capable validators should expose what they checked, which version they checked, and what remains unknown. A human reviewer could then assess a concrete proposed recovery point and action rather than approve a vague instruction to continue. Human authorization and reconciliation of conflicting evidence remain future work.

\subsection{Checking Both Execution and Its Explanation}
A recovery decision is useful only if it reaches the executor. The implementation must identify the path through which an allowed action obtains permission and what prevents that same path from being used after withholding. Recovery limits belong on this path as well: an agent that repeatedly requests recovery can still fail to make progress.

Checking the resulting record raises a separate question: who controls the expected policy? Study D shows that a client can alter both execution and its reported reference, producing internal agreement that is insufficient to establish conformance. Keeping the reference independently lets the system reject that explanation even after an effect occurs. This is an audit acceptance property, distinct from stopping the effect in advance.

Together with the source-\hspace{0pt}selection studies, this yields two distinct lessons. Accurate restoration and final success do not establish that the source decision was permissible; an observed effect and internally consistent records do not establish agreement with the expected policy. Recovery records should therefore expose the selected source, route, reason, execution evidence, and basis for accepting the run. The experiments establish these distinctions within their supplied contracts and, for the authority apparatus, its trusted caller and process boundary. They do not establish general isolation from a compromised host.

\subsection{Evaluating Recovery Mechanisms}
A detailed EvidencePackage is useful for inspection, but its size does not measure the quality of the decision. In the current grant-\hspace{0pt}path router, the failure family and recorded eligible checkpoint determine the choice; other retained fields have no demonstrated causal benefit.

To evaluate richer evidence, future cases should hold the apparent failure fixed while changing something that matters to recovery. The same timeout could permit retry when no effect occurred and require withholding when an external write is unresolved. A document checkpoint could be reusable under a current validation and ineligible after that evidence expires. These comparisons would test whether the system changes the source or route for the relevant reason. Removing unused fields from the present implementation would not provide that evidence.

Comparisons with semantic checkpoint managers and workflows should supply the same candidates, validators, event evidence, and repair resources. Holding routes fixed while varying source rules would isolate source selection; holding sources fixed while varying route policies would examine routing. Both required continuation and required withholding should be represented, with decision correctness, enactment, restoration fidelity where applicable, and final outcomes reported separately. Systems that satisfy the same contract could then be compared on retained work and the costs of validation, storage, inference, and enforcement. This distinguishes implementing the responsibility from implementing it efficiently; the present experiments do not provide that complete comparison.

\section{Limitations and Future Work}
\subsection{Given Rules and Narrow Task Coverage}
The studies supply the validators and continuation policies. They test whether the runtimes follow these rules, not whether the rules correctly capture every real-\hspace{0pt}world condition. A passed validator is evidence within its scope, not proof that a state is objectively trustworthy. The work does not evaluate who should define or update the policy, or how disagreements among validators should be resolved. Policy discovery and assessment under uncertain evidence remain unresolved.

Studies A and B use scripted tasks, failures, and repairs. Their failure detectors recognize deliberately created symptoms. Study B changes prefix length with a checkpoint after every validated step, rather than reproducing the broader challenges of long-\hspace{0pt}horizon execution. Study C adds model-\hspace{0pt}generated text and real file restoration, but uses four invariant templates, one local Markdown file, one write-\hspace{0pt}then-\hspace{0pt}timeout pattern, and one endpoint-\hspace{0pt}reported model ID. Its selected challenge candidates do not estimate natural failure rates. Study D uses 18 matched deterministic cases and synthetic probes. None of these matrices is a representative deployment sample or evidence of cross-\hspace{0pt}model robustness.

\subsection{Comparisons and Costs Not Yet Measured}
The restoring modes share storage and restoration, but source and fixed route rules change together. The selected source is directly observable; the separate contribution of route selection to outcomes has not been isolated. Stronger source policies, semantic checkpoint managers, workflow engines with comparable controls, and human escalation remain important untested comparisons.

The progress result counts actions under unit checkpoint spacing. It does not measure the costs of model inference, validation, checkpoint I/\hspace{0pt}O, storage, audit, or permission enforcement. Wider checkpoint intervals, different action costs, and irreversible effects could change the tradeoff. Study C records latency and tokens, but its modes repair different restored inputs, so those records do not establish a controlled efficiency comparison.

\subsection{Trust and Reproduction Boundaries}
The grant-\hspace{0pt}path studies do not independently test permission at the failure event or exercise required refusal. They also do not evaluate adversarial modification of evidence, checkpoint pointers, or decision records. Study D adds event-\hspace{0pt}time decisions and adversarial cases, but assumes integrity of the trusted caller and spawn mechanism under its declared apparatus-\hspace{0pt}channel adversary. It does not cover OS compromise, arbitrary external effects, or replacement of the trusted configuration builder. Rejecting the adversarial execution as nonconforming also does not prevent or reverse its physical effect.

Offline verification checks retained evidence within the artifact's scope. It cannot recreate the original model-\hspace{0pt}serving environment or re-\hspace{0pt}observe the OS exit status of an already exited process. The external action ledger prevents task rollback from erasing incurred events in the implementation; its resistance to adversarial tampering has not been evaluated. Appendix A retains the additional evidence and reproduction qualifications.

\subsection{Extending the Decision beyond the Current Cases}
Cross-\hspace{0pt}session handoff, memory conflict resolution, and partial rollback across dependent artifacts remain design extensions. The experiments do not show autonomous repair quality, production reliability, or a benefit from every field in the evidence representation.

Future evaluations should combine multiple artifacts, changing evidence, and compound failures with recovery actions whose validity depends on those conditions. They should examine diagnosis, evidence assessment, source choice, route choice, repair generation, and verification separately, while also measuring the whole control path. This would test whether the contract remains useful as tasks and evidence become less constructed.

\section{Conclusion}
When an agent's work is interrupted or stops making reliable progress, recovery must decide what can be retained and how to continue. We introduced recoverability as a system primitive that makes reuse an explicit, checkable decision: a permitted source and action, or required withholding. Its behavioral contract connects that decision to supporting evidence, execution, and recovery bounds. The RPR architecture shows how persistence, validation, planning, and control can implement this responsibility.

The controlled studies establish why this decision needs its own evaluation. Accurate restoration and eventual success can conceal an excluded starting point. An observed effect and internally consistent records can still violate the independently specified policy. Explicit eligibility rules address the first problem in the tested cases; independently held policy evidence exposes the second. Shared checkpoint restoration supplies the demonstrated benefit of retaining validated work. These observations distinguish preservation, permission, execution, and acceptance without claiming a general reliability advantage over complete recovery systems.

Within supplied policies and the declared trust model, the framework turns conditions for continuing into requirements that implementations can enforce and evaluators can test. It does not infer trustworthy policies in an open environment, and detecting a violation after an effect does not undo it. Its contribution is to make the basis for reusing progress independently specifiable and checkable, including when eventual task success would conceal a recovery error.

\label{page:main-end}
\clearpage
\bibliographystyle{plainnat}
\bibliography{references}
\clearpage
\appendix
\section{Reproducibility and Evidence Scope}
\subsection{Frozen Implementations and Model Calls}
The deterministic runtime and file runtime are frozen as separate artifacts. RPR-\hspace{0pt}Grant uses tag \code{cx01-\hspace{0pt}v03-\hspace{0pt}phase2-\hspace{0pt}common-\hspace{0pt}cost}, commit \code{1a4f\hspace{0pt}3851\hspace{0pt}a266\hspace{0pt}2541\hspace{0pt}5d97\hspace{0pt}5e51\hspace{0pt}98ed\hspace{0pt}70ad\hspace{0pt}6106\hspace{0pt}a012}; RPR-\hspace{0pt}File uses tag \code{cx01-\hspace{0pt}v04-\hspace{0pt}phase4-\hspace{0pt}real-\hspace{0pt}case}, commit \code{cb8a\hspace{0pt}ab7d\hspace{0pt}fb25\hspace{0pt}4250\hspace{0pt}6b08\hspace{0pt}f3a3\hspace{0pt}55bb\hspace{0pt}6a66\hspace{0pt}a3d2\hspace{0pt}fcd8}. Each includes a \code{FREEZE\_\hspace{0pt}MANIFEST.md} binding its retained evidence. The implementations and evidence are retained in a private versioned release candidate. Their presence in that release should not be read as public availability with this manuscript. UUID-\hspace{0pt}derived object identifiers can vary in a fresh deterministic run, so reproduction targets the reported source, fidelity, invariant, and accounting results rather than identical raw JSONL bytes.

Study C used \code{openai==2.45.0}, Responses structured parsing with a two-\hspace{0pt}field document schema, reasoning effort \code{low}, \code{max\_\hspace{0pt}output\_\hspace{0pt}tokens=500}, \code{store: false}, a 90-\hspace{0pt}second timeout, and at most two SDK transport retries. Calls used the OpenAI-\hspace{0pt}compatible endpoint \code{coder.api.visioncoder.cn}, which reported model ID \code{gpt-\hspace{0pt}5.6-\hspace{0pt}luna}; the upstream model and infrastructure were not independently verified. Transport retries are distinct from the content-\hspace{0pt}level allowance of one additional candidate-\hspace{0pt}generation attempt. Only retained challenge candidates are serialized. The record contains 20 candidate and 40 repair generations, with no observed error in that retained matrix; SDK request IDs are \code{null} in all 60 records. This is not a count of all provider calls. Offline verification checks the retained code, outputs, and 40 workspaces; repeating the external calls depends on endpoint access and a potentially changed serving environment.

Study D's results are bound to candidate tree v14 and \code{runs/\hspace{0pt}g2de-\hspace{0pt}r13-\hspace{0pt}2026-\hspace{0pt}08-\hspace{0pt}22/\hspace{0pt}}, with 26 frozen contracts. The separate archive \code{study\_\hspace{0pt}artifact\_\hspace{0pt}d1b7\hspace{0pt}d4bc.zip} has SHA-256 \code{d1b7\hspace{0pt}d4bc\hspace{0pt}9de2\hspace{0pt}4b7d\hspace{0pt}cad4\hspace{0pt}3b5f\hspace{0pt}9d04\hspace{0pt}be74\hspace{0pt}3b60\hspace{0pt}5b98\hspace{0pt}62f8\hspace{0pt}8d8a\hspace{0pt}ba53\hspace{0pt}7fbf\hspace{0pt}360d\hspace{0pt}148e}. It vendors the required base and provides an offline verification path. Later packaging changes through the sealed v20 workspace leave the experimental \code{studyd/\hspace{0pt}} implementation unchanged. The artifact supports recomputation of retained deterministic evidence under the declared profile.

\subsection{Qualifications of the Authority Evidence}
The retained Study D record has six specific qualifications:

\begin{enumerate}
  \item \textbf{Enforcement row binding.} The 84 synthetic enforcement probes do not materialize an 84-\hspace{0pt}row case-\hspace{0pt}bound manifest with per-\hspace{0pt}row \code{row\_\hspace{0pt}sha256} bindings. Their structure and pass conditions are fixed by frozen manifest v1; observations are recomputed from sealed authority sessions.
  \item \textbf{Evidence shapes.} The 30 semantic controls each retain three packet/\hspace{0pt}trace documents and independent-\hspace{0pt}process oracle output. Enforcement and recovery-\hspace{0pt}bound probes instead retain per-\hspace{0pt}session authority evidence with a hash-\hspace{0pt}chained transcript and replayed verification; they do not claim the same packet shape.
  \item \textbf{Threat and integrity boundaries.} The apparatus claim uses the threat model in Section 5.3. Post hoc retained-\hspace{0pt}byte integrity is checked separately through the run manifest, tree and Git identities, and governance records.
  \item \textbf{Archive completeness.} The sealed workspace alone depends on a commit-\hspace{0pt}pinned public base. The vendored archive supplies the self-\hspace{0pt}contained offline form.
  \item \textbf{Historical records.} Date errors in historical governance records are corrected through later clarification entries, preserving the earlier recorded hashes. Terminal evidence identities govern the results reported here.
  \item \textbf{Exited processes.} An offline auditor cannot re-\hspace{0pt}observe an exited process's OS exit code. It verifies the seal's internal \code{aborted\_\hspace{0pt}by}/\hspace{0pt}exit-\hspace{0pt}code mapping; the live runner/\hspace{0pt}client separately observes the process-\hspace{0pt}exit interlock.
\end{enumerate}

These qualifications distinguish what the retained bytes permit an auditor to recompute from observations that require the original live execution or an additional trust assumption.

\subsection{Authority Verification Inventory and Comparison Identifiers}
The independent offline pass recomputed all 392 files in the run manifest, verified 210 authority evidence documents, and reconciled the 54 qualification/\hspace{0pt}main records with zero layer-4 failures. These inventory counts describe retained verification coverage, not additional independent task samples.

In the adversarial comparison in Section 6.5, the replacement client expands the handlers permitted for \code{resume\_\hspace{0pt}retry} to include \code{cursor\_\hspace{0pt}state\_\hspace{0pt}sink} and causes the real authority process to advance its cursor from 0 to 1. Round 12 copies its expected reference from \code{client.config} after construction and accepts the internally consistent record as \code{ok}. Round 13 retains the caller's canonical configuration before client construction and rejects conformance as \code{mismatch/\hspace{0pt}authority\_\hspace{0pt}evidence}; offline verification confirms this rejection. Both conditions produce the physical effect. The change concerns acceptance against independently held policy evidence, not prevention or rollback.

\label{page:appendix-end}
\end{document}